\documentclass[runningheads]{llncs}
\usepackage[T1]{fontenc}
\usepackage{graphicx}
\usepackage{booktabs} % For formal tables
\usepackage{multirow} % For multirow cells
\usepackage{arydshln}
\usepackage[colorlinks=true, linkcolor=blue, citecolor=blue, urlcolor=blue]{hyperref}
\usepackage{amssymb}
\usepackage{amsmath}
\usepackage{caption}
\usepackage{marvosym}

\begin{document}
\title{SCMIL: Sparse Context-aware Multiple Instance Learning for Predicting Cancer Survival Probability Distribution in Whole Slide Images}
\titlerunning{Sparse Context-aware Multiple Instance Learning}
% If the paper title is too long for the running head, you can set
% an abbreviated paper title here
%
\author{Zekang Yang\inst{1,2} \and Hong Liu\inst{1} $^{(\textrm{\Letter})}$ \and Xiangdong Wang\inst{1}}
\authorrunning{Z. Yang et al.}
% First names are abbreviated in the running head.
% If there are more than two authors, 'et al.' is used.
%
\institute{Beijing Key Laboratory of Mobile Computing and Pervasive Device, Institute of Computing Technology, Chinese Academy of Sciences, Beijing 100190, China. 
\email{hliu@ict.ac.cn}\\
\and University of Chinese Academy of Sciences, Beijing 100086, China.
}
% index{Yang, Zekang}
% index{Liu, Hong}
% index{Wang, Xiangdong}

%
\maketitle              % typeset the header of the contribution
\begin{abstract}
Cancer survival prediction is a challenging task that involves analyzing of the tumor microenvironment within Whole Slide Image (WSI).
Previous methods cannot effectively capture the intricate interaction features among instances within the local area of WSI.
Moreover, existing methods for cancer survival prediction based on WSI often fail to provide better clinically meaningful predictions.
To overcome these challenges, we propose a Sparse Context-aware Multiple Instance Learning (SCMIL) framework for predicting cancer survival probability distributions.
SCMIL innovatively segments patches into various clusters based on their morphological features and spatial location information, subsequently leveraging sparse self-attention to discern the relationships between these patches with a context-aware perspective.
Considering many patches are irrelevant to the task, we introduce a learnable patch filtering module called SoftFilter, which ensures that only interactions between task-relevant patches are considered.
To enhance the clinical relevance of our prediction, we propose a register-based mixture density network to forecast the survival probability distribution for individual patients.
We evaluate SCMIL on two public WSI datasets from the The Cancer Genome Atlas (TCGA) specifically focusing on lung adenocarcinom (LUAD) and kidney
renal clear cell carcinoma (KIRC).
Our experimental results indicate that SCMIL outperforms current state-of-the-art methods for survival prediction, offering more clinically meaningful and interpretable outcomes.
Our code is accessible at \url{https://github.com/yang-ze-kang/SCMIL}.

% The abstract should briefly summarize the contents of the paper in 150--250 words.

\end{abstract}
\section{Introduction}
Using Whole Slide Image (WSI) to predict patient’s cancer survival risk is crucial for health monitoring and personalized treatment in clinical settings.
Pathologists typically examine WSIs manually to identify relevant biological features for diagnosis.
However, the high resolution of WSI demands considering time and effort to complete the analysis.
Automatic diagnosis using deep learning technology has the potential to significantly reduce the workload of pathologists, and many studies have been conducted on this subject \cite{chen2022pan,clam,yao2020whole}.
% Automatic diagnosis using deep learning technology has the potential to significantly reduce the workload of pathologists, and many studies have been conducted on this subject \cite{chen2022pan,clam}.
% Due to the high resolution of WSIs, there are GPU memory limitations challenges.
Obtaining fine-grained annotations for high-resolution WSI is challenging, and it is often treated as a weakly supervised learning task.
In recent years, researchers have developed various methods to address this challenge, achieving commendable results in cancer diagnosis.
Unlike cancer diagnosis, survival risk prediction involves not only extracting biomorphological features but also delving into the interactions between cells and tissues within the tumor microenvironment.
Furthermore, providing predictions with enhanced clinical relevance posed an additional challenge in the task of survival prediction \cite{haider2020effective}.

Due to the high resolution of WSIs, it is common practice to segment them into patches with a fixed size.
Then a feature extractor, such as ImageNet pretrained ResNet50 \cite{he2016deep}, is used to extract features from all patches, followed by multiple instance learning \cite{amil} for predictive analysis.
Methods like AMIL \cite{amil}, CLAM \cite{clam}, and DSMIL \cite{dsmil} make predictions by identifying key patches.
However, these methods neglect the interaction among patches, which is insufficient for survival prediction tasks.
Approaches such as WSISA \cite{zhu2017wsisa}, and DeepAttnMISL \cite{yao2019deep} use clustering to divide patches into various phenotypes and then extract the features of each phenotype respectively.
While these methods consider the morphological relationship between patches, they disregard the spatial connections.
Methods like PatchGCN \cite{patchgcn}, and HGT \cite{hgt} treat WSIs as point clouds with each patch represented as a node. Graph Convolutional Networks (GCNs) \cite{hamilton2017inductive,gcn,xu2018powerful} are used to explore the relationships among patches.
In these methods, each patch pays attention to the information from neighboring patches, requiring deeper layers to cover a wider area.
However, an increase in layer depth leads to a significant rise in computational demands and GPU memory usage. And the mining of the relationship among patches also depends on the selection of aggregation function.
TransMIL \cite{transmil} employs a self-attention mechanism along with the PPEG module to investigate inter-patch relationships. However, to mitigate GPU memory constraints, the author uses linear approximation for self-attention, resulting in a coarse-grained attention between patches.

To address the aforementioned challenges, we propose a Sparse Context-aware Multiple Instance Learning (SCMIL) framework for the prediction of patient survival probability distributions.
Our primary contributions are as follows:
(1) We design a patch filtering module called SoftFilter to identify task-relevant patches and can be trained through backpropagation.
(2) We propose the Sparse Context-aware Self-Attention (SCSA), which uses sparse self-attention to learn the interactions among local patches, while concurrently incorporating both spatial and morphological information to guide the learning of patch interactions in specific areas.
(3) We present the Register-based Mixture Density Network (RegisterMDN), which can learn the parameters for each component of a Gaussian Mixture Model from data of cancer patient cohort and utilizes individual patient's data to forecast the weights of these components. This approach enables the prediction of a tailored survival probability curve for each patient and enhances the interpretability and clinical significance of the model’s predictions.

\section{Methodology}
\begin{figure*}[t]
  \centering
    \includegraphics[width=\textwidth]{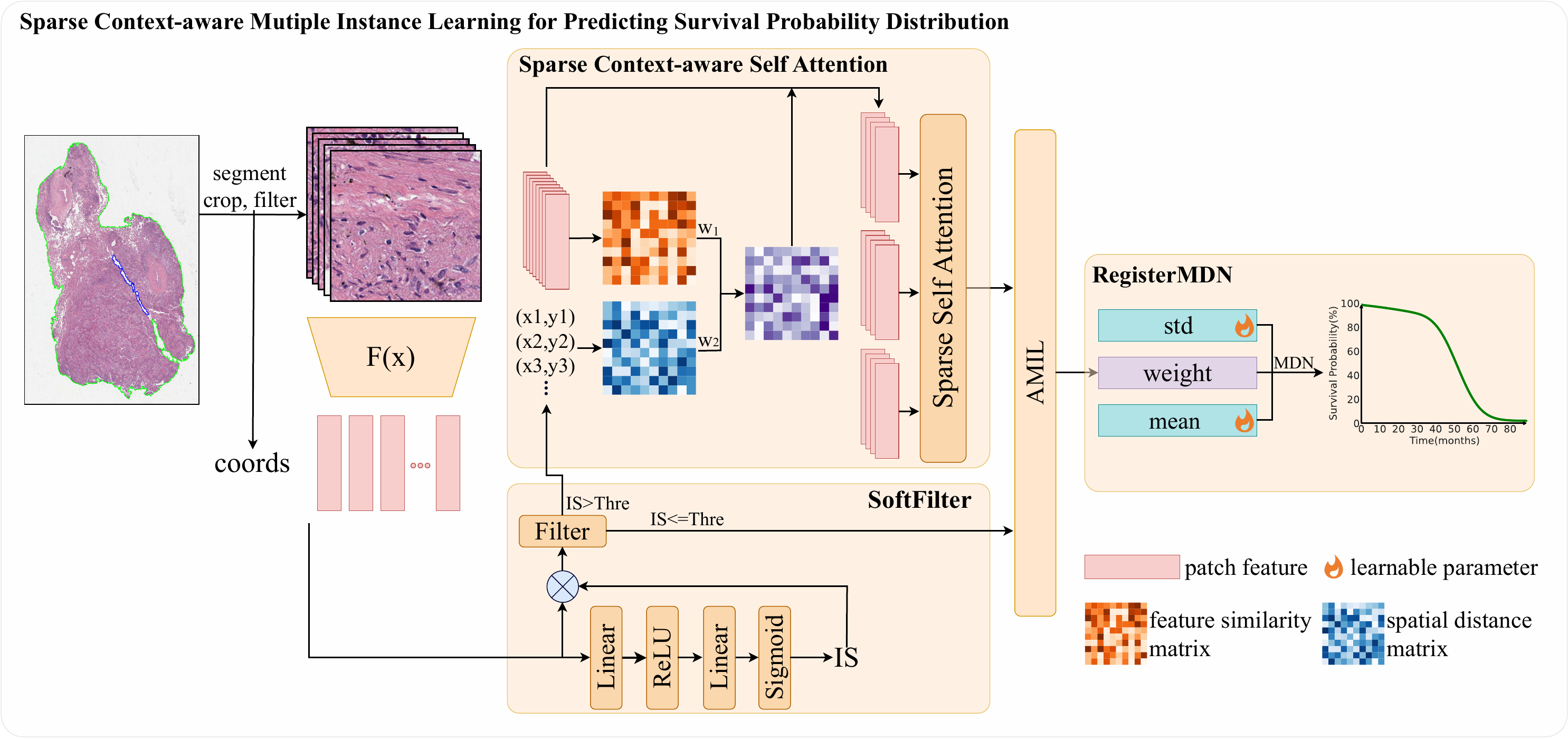}
  \caption{Overview of the proposed Sparse Context-aware Multiple Instance Learning (SCMIL) framework for predicting cancer survival probability distribution.} 
  \label{fig:overview}
\end{figure*}

Figure \ref{fig:overview} depicts the pipeline of our proposed Sparse Context-aware Multiple Instance Learning (SCMIL) framework.
WSIs are segmented into fixed-size patches with 256×256 pixels, and irrelevant patches are filtered out.
Subsequently, we use the feature extractor ViT \cite{vit} ($F(x)$ in Figure \ref{fig:overview}), which has been pre-trained on a large-scale collection of WSIs using self-supervised learning \cite{kang2023benchmarking}, to extract the features $Feat \in \mathbb{R}^{n\times d}$ for all patches.
The fundamental principle  of our SCMIL approach is to identify regions within high-resolution WSI that are most informative for predicting patient survival risk.
In these significant areas, we identify biomarkers that are associated with survival risk.
By integrating the survival information from the cancer patient cohort, we can subsequently generate a survival probability distribution for the patient.
SCMIL framework is mainly composed of three components: SoftFilter, Sparse Context-aware Self-Attention (SCSA), and the Register-based Mixture Density Network (RegisterMDN).
SoftFilter help SCSA focus on task-specific areas, and RegisterMDN predicts the survival probability distribution based on the wsi-level feature.
\subsection{SoftFilter}
Within each WSI, there exist numerous patches that are irrelevant to the immediate task. To address this problem, we design a learnable patch filtering module termed SoftFilter.
SoftFilter inputs the features of patches into a Multilayer Perceptron (MLP) followed by a Sigmoid activation function to predict the patches' importance scores $IS \in \mathbb{R}^{n\times1}$:
\begin{equation}
    IS = Sigmoid(MLP(Feat))
\end{equation}
Subsequently, the features of each patch are element-wise multiplied by their corresponding importance score to derive the new features $H\in\mathbb{R}^{n\times d}$.
This process enables the SoftFilter module learnable without requiring patch-level supervision.
$H$ are then partitioned into task-relevant features $H_{high}$ and task-irrelevant features $H_{low}$ according to the IS threshold $Thre$.
The task-relevant features are propagated to the SCSA module for learning the interactions among patches, while the task-irrelevant features bypass this stage.

\subsection{Sparse Context-aware Self-Attention (SCSA)}
After obtaining the task-relevant features, we devise a Sparse Context-aware Self-Attention (SCSA) module to explore the interactions among patches.
The SCSA first cluster the potentially interacting patches into the $C$ clusters $\{L_1, \allowbreak L_2, ..., L_C\}$ based on the morphological features and spatial positions of the patches.
Specifically, we employ the K-Means clustering algorithm to divide the task-relevant patches and the similarity between patches is obtained by a weighted sum of the cosine similarity of morphological features and the normalized Euclidean distance of spatial positions, with the weights being $w_1$ and $w_2$ respectively.
To accommodate WSIs of varying sizes, we fix the size of the clusters and derive the number of clusters from the size of the clusters.
Then we utilize the Multi-Head Self-Attention mechanism (MHSA) \cite{vaswani2017attention} to learn the relationships within each cluster and obtain refined features $L_i'$
\begin{equation}
    L_{i}' = MHSA(L_i)+L_i,i=1,2,...,C
\end{equation}
Compared with linear self-attentions methods \cite{transmil,xiong2021nystromformer}, our sparse self-attention approach enables a more fine-grained attention to the relationships among patches.
Subsequently, the features from all clusters, along with the task-irrelevant features, are concatenated. The WSI-level features $Feat'$ is obtained through an attention-weighted process \cite{amil}:
\begin{equation}
    H' = Concat(L_1',L_2',...,L_C',H_{low})
\end{equation}
\begin{equation}
    \alpha_{i} = \frac{exp(a^T(tanh(V H_{i}'^T)\odot \sigma(UH_{i}'^T)))}{\sum_{k=1}^{n}exp(a^T(tanh(VH_{k}'^T)\odot \sigma(UH_{k}'^T)))}
\end{equation}
\begin{equation}
    Feat' = \sum_{i=1}^{n}\alpha_i H_{i}'
\end{equation}
where $U$, $V$, and $a$ are learnable parameters, $n$ is the number of patches within the WSI, $\odot$ denotes element-wise multiplication, and $tanh()$ is the hyperbolic tangent function.
The features $Feat'$ now contain biomarkers relevant to patient survival risk and are instrumental in subsequent survival prediction tasks.

\subsection{RegisterMDN}
Previous studies \cite{hipt,patchgcn,hgt,yao2019deep} for predicting survival risk based on WSIs mainly focus on predicting a time-independent risk value.
This approach is of limited utility when considering only the risk value of an individual patient.
A more comprehensive prognosis of a patient’s survival risk should take into account the risk values and survival times of other patients within the cancer patient cohort.
Moreover, looking at the risk value for a single patient does not provide useful information.
To provide more clinically meaningful predictions, we design the Register-based Mixture Density Network (RegisterMDN) inspired by SurvivlMDN \cite{han2022survival} to predict the survival probability distribution for an individual patient.

The Mixed Density Network (MDN) translates the input to a probability distribution.
We adopt Gaussian distributions as the components of the MDN, assuming that the number of components is $K$.
We utilize the WSI-level features $Feat'$, the mean vector $P_{m}$, and the standard deviation vector $P_{v}$ as the input of our RegisterMDN.
Both $P_{m}$ and $P_{v}$ are learnable parameters and learn the survival risk characteristics of the specific cancer during the training phase.
$Feat'$, $P_{m}$, and $P_{var}$ through the neural networks to produce the weights $\lambda_i(Feat')$, means $\mu_i(P_{m})$, and variances $\sigma_i^2(P_{v})$ of the mixture model.
Consequently, we can get the Probability Density Function (PDF):
\begin{equation}
    PDF(y|Feat',P_{m},P_{v}) = \sum_{i=1}^K\lambda_i(Feat')\mathcal{N}(y|\mu_i(P_{m}),\sigma_i^2(P_{v}))
\end{equation}
The patient's survival time is a positive number, so we define the survival time $t=g(x)=log(1+exp(y))$. This transformation enables us to formulate the patient’s Death Probability Density Function (DPDF) and Death Cumulative Density Function (DCDF):
\begin{equation}
    DPDF(t|Feat',P_{m},P_{v}) = |\frac{\mathrm{d}g^{-1}}{\mathrm{d}t}|\sum_{i=1]}^K\lambda_i(Feat')\mathcal{N}(g^{-1}(t)|\mu_i(P_{m}),\sigma_i^2(P_{v}))
\end{equation}
\begin{equation}
    DCDF(t|Feat',P_{m},P_{v}) = \sum_i^{K}\lambda_i(Feat')\mathrm{erf}(\frac{g^{-1}(t)-\mu_i(x)}{\sigma_i(x)})
\end{equation}
where erf($\cdot$) is the Gaussian error function.
The patient's Survival Cumulative Distribution Function $SCDF(t|Feat',P_{m},P_{v}) = 1 - DCDF(t|Feat',P_{m},P_{v})$ is the final predicted patient survival probability distribution.

Assuming the patient's right uncensorship status is c (1 for uncensored data and 0 for censored data), the duration from diagnosis to death is $d$, and the time from diagnosis to the last follow up is $o$. $td$ is either equal to $d$ ($c=1$) or $o$ ($c=0$). Then we can define the loss function of RegisterMDN with the help of maximum likelihood estimation:
\begin{equation}
\begin{aligned}
    loss &= -c\cdot\mathrm{log}(DPDF(td|Feat',P_m,P_v)) \\
        &-(1-c)\cdot\mathrm{log}(SCDF(td|Feat',P_m,P_v))
\end{aligned}
\end{equation}
\section{Experiments}
\subsection{Experimental Settings}
\subsubsection{Dataset.}
We evaluate the effectiveness of our method on The Caner Genome Atlas (TCGA) lung adenocarcinom (LUAD) with 452 cases and kidney renal clear cell carcinoma (KIRC) with 512 cases.
All WSIs are analyzed at 20x magnification and cropped into 256 × 256 patches.
The average number of patches per WSI is 12,097 for TCGA-LUAD, and 14,249 for TCGA-KIRC, with the largest number of patches is 84,365 from a TCGA-KIRC sample.
\subsubsection{Implementation Details.}
In our implementation, we set the cluster size $C$ in SCSA to be 64, the threshold $Thres$ in SoftFilter to be 0.5, and the number of components $K$ in RegisterMDN to be 100.
We use cuML \cite{raschka2020machine} to accelerate the execution of the K-Means algorithm on the GPU.
For all comparison experiments and ablation experiments, we maintain a consistent hyperparameter setting: the learning rate of 2e-4 with a weight decay of 1e-3, the Adam optimizer is used to update the model weights, a dropout rate of 0.1, a batch size of 1, and training for 20 epochs.
The 5-fold cross-validation are used on all datasets and models.
\subsubsection{Evaluation Metric.}
The conventional concordance index (C-Index) \cite{wang2019machine} is limted to provide a more comprehensive comparison between different methods.
We introduce enhanced evaluation metrics.
We use a time-dependent version of the concordance estimator (TDC) within a pre-specified time span $[0,\tau]$.
TDC measures the proportion of patients pairs for which the survival risks is correctly ranked at multiple time points in $[0,\tau]$.
The Brier score (BS) calculates the mean square error between the ground-truth and the predicted probability.
It mainly measures the calibration performance. To consider all times, we use an integrated BS (IBS) over time interval $[0,\tau]$.
Models with larger TDC and lower IBS demonstrate superior performance.
The result of mean ± std is reported.
\setlength\tabcolsep{4pt}
\begin{table}[htbp]
    \centering
    \caption{Evaluation of all models on TCGA-KIRC and TCGA-LUAD with time dependent concordance index (TDC) and integrated Brier Score (IBS). Best results are marked in bold.}
    \label{tab:model_comparison}
    \resizebox{0.98\textwidth}{!}{
        \begin{tabular}{cccccc}
            \toprule
            \multirow{2}{*}{Method} & \multicolumn{2}{c}{KIRC} & \phantom{a} & \multicolumn{2}{c}{LUAD} \\
            \cmidrule{2-3} \cmidrule{5-6}
                                            & TDC\,$\uparrow$  & IBS\,$\downarrow$         &&   TDC\,$\uparrow$ & IBS\,$\downarrow$          \\ 
            \midrule
            AMIL \cite{amil}                & 0.627\,$\pm$\,0.063  & 0.287\,$\pm$\,0.014     && 0.612\,$\pm$\,0.042  & 0.305\,$\pm$\,0.045   \\
            % DeepAttnMISL \cite{yao2019deep} & \,$\pm$\,  & \,$\pm$\,   && \,$\pm$\,   & \,$\pm$\,        \\
            CLAM \cite{clam}                & 0.664\,$\pm$\,0.037  & 0.289\,$\pm$\,0.031     && 0.592\,$\pm$\,0.070  & 0.308\,$\pm$\,0.044   \\
            DSMIL \cite{dsmil}              & 0.642\,$\pm$\,0.045  & 0.289\,$\pm$\,0.015     && 0.581\,$\pm$\,0.075  & 0.322\,$\pm$\,0.044   \\
            PatchGCN \cite{patchgcn}        & 0.674\,$\pm$\,0.049  & 0.279\,$\pm$\,0.026     && 0.582\,$\pm$\,0.055  & 0.307\,$\pm$\,0.045   \\
            TransMIL \cite{transmil}        & 0.629\,$\pm$\,0.041  & 0.290\,$\pm$\,0.017     && 0.512\,$\pm$\,0.082  & 0.319\,$\pm$\,0.033   \\
            HIPT \cite{hipt}                & 0.635\,$\pm$\,0.041  & 0.270\,$\pm$\,0.021     && 0.540\,$\pm$\,0.025  & 0.289\,$\pm$\,0.068        \\
            HGT \cite{hgt}                  & 0.634\,$\pm$\,0.058  & 0.269\,$\pm$\,0.033     && 0.601\,$\pm$\,0.042  & 0.289\,$\pm$\,0.052        \\
            \hdashline
            SCMIL w/o SoftFilter            & 0.659\,$\pm$\,0.038  & 0.278\,$\pm$\,0.015     && 0.546\,$\pm$\,0.046  & 0.318\,$\pm$\,0.043   \\
            SCMIL w/o SCSA                  & 0.651\,$\pm$\,0.020  & 0.274\,$\pm$\,0.015     && 0.589\,$\pm$\,0.042  & 0.318\,$\pm$\,0.028   \\
            \textbf{SCMIL}                  & \textbf{0.688\,$\pm$\,0.037}  & \textbf{0.268\,$\pm$\,0.021}    && \textbf{0.622\,$\pm$\,0.015}  & \textbf{0.288\,$\pm$\,0.060}  \\
            \bottomrule
        \end{tabular}
    }
\end{table}
\subsection{Experiments and Results}
\subsubsection{Comparison with State-of-the-Art Methods.}
To compare the ability of our proposed SCMIL in learning cancer survival risk-related features with existing methods, we select several state-of-the-art methods, including AMIL \cite{amil}, CLAM \cite{clam}, DSMIL \cite{dsmil}, PatchGCN \cite{patchgcn}, TransMIL \cite{transmil}, HIPT \cite{hipt}, and HGT \cite{hgt}.
We add the RegisterMDN module into these methods to predict the patient's survival probability distribution, ensuring a fair comparison with our method.
SCMIL demonstrates its ability to learn interactions between related patches, which is an advancement over methods based on key patches \cite{amil,clam,dsmil}.
Compared to GCNs-based methods \cite{patchgcn,hgt} that focuses on adjacent patches, SCMIL offer a more adaptable attention scope.
SCMIL also outperforms Transformer-based methods \cite{hipt,transmil} that emphasize global patches by focusing more effectively on local regions of interest.
The experimental results are presented in Table \ref{tab:model_comparison}.
Our proposed SCMIL has achieved the best performance in both TDC and IBS metrics on two WSI datasets, proving its superior ability to learn features associated with cancer survival risk from WSIs compared to previous methods.

\subsubsection{Ablation Analysis.}
Table \ref{tab:model_comparison} presents the experimental results on SCMIL with the removal of the SoftFilter module and the SCSA module, respectively.
The omission of either module lead to a decline in performance, underscoring the essential role of both modules.
Notably, the model’s performance on the LUAD dataset is significantly decreased without the SoftFilter module, which suggests that many patches in this dataset may be irrelevant to the task.
\begin{figure}[bp]
  \centering
  \begin{minipage}[c]{0.4\linewidth}
    \centering
    \includegraphics[width=\linewidth]{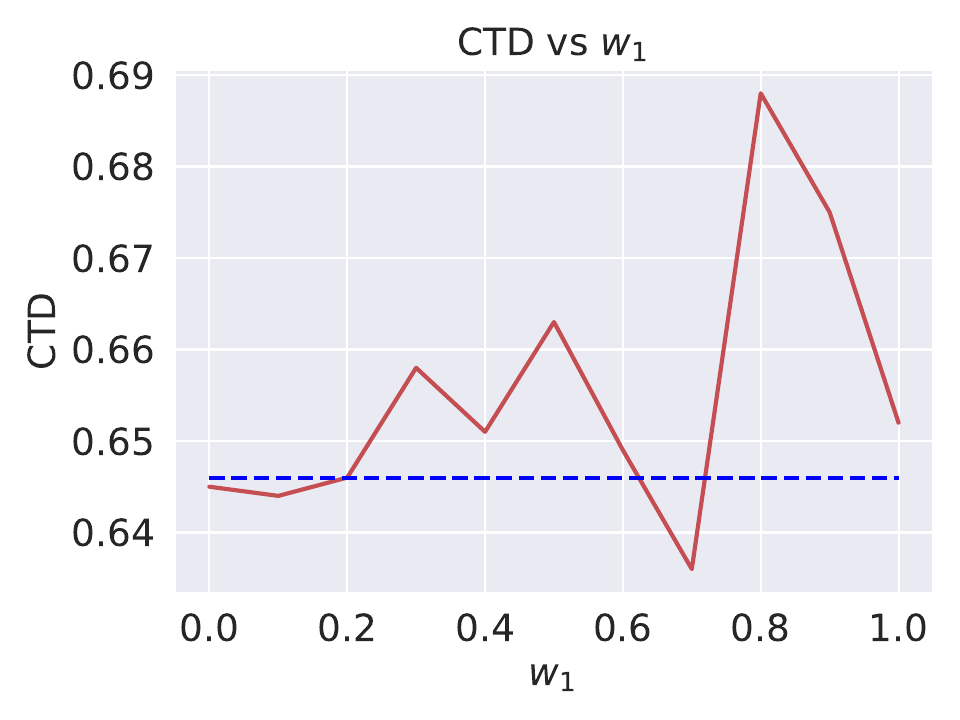}
    \captionof{figure}{Comparison of different clustering methods.}
    \label{fig:weight}
  \end{minipage}
  \begin{minipage}[c]{0.57\linewidth}
    \centering
    \captionof{table}{Comparison of different probability distribution prediction methods. Best results are marked in bold, second best results are underlined.}
    \resizebox{\textwidth}{!}{
    \begin{tabular}{ccc}
        \hline
        Method  & TDC\,$\uparrow$  & IBS\,$\downarrow$  \\
        \hline
        Predicted Vector \cite{han2022survival} & 0.653\,$\pm$\,0.100  &\textbf{0.255\,$\pm$\,0.017}   \\
        Fixed Vector & \underline{0.683\,$\pm$\,0.018}  &0.280\,$\pm$\,0.023   \\
        Learnable Vector & \textbf{0.688\,$\pm$\,0.037}  & \underline{0.268\,$\pm$\,0.021}   \\
        \hline
    \end{tabular}}
    \label{tab:RegisterMDN}
  \end{minipage}
\end{figure}
Further experiments on the KIRC dataset are conducted to assess the impact of varying morphological similarity weight $w_1$ and spatial location similarity weight $(1-w_1)$ on model performance during clustering.
Figure \ref{fig:weight} illustrates these experimental results, with the blue dotted line indicating the experimental results from random clustering.
It is evident that an 8:2 weighted ratio of morphological similarity to spatial location similarity yields the best model performance.
Conversely, models that rely solely on morphological information or spatial location information for clustering exhibit inferior performance.
We further evaluate various approaches for predicting the survival probability distribution: (1) Predicted Vector, which forecasts the parameters of each MDN component via $Feat'$; (2) Fixed Vector, which predefines the parameters of each component in advance; (3) Learnable Vector, a method we designed that allows for learning parameters. The experimental results, as shown in Table \ref{tab:RegisterMDN}, indicating that our proposed Learnable Vector method offers superior discriminative power and improved calibration.
\begin{figure}[t]
  \centering
  \begin{minipage}[c]{0.49\linewidth}
    \centering
    \includegraphics[width=\linewidth]{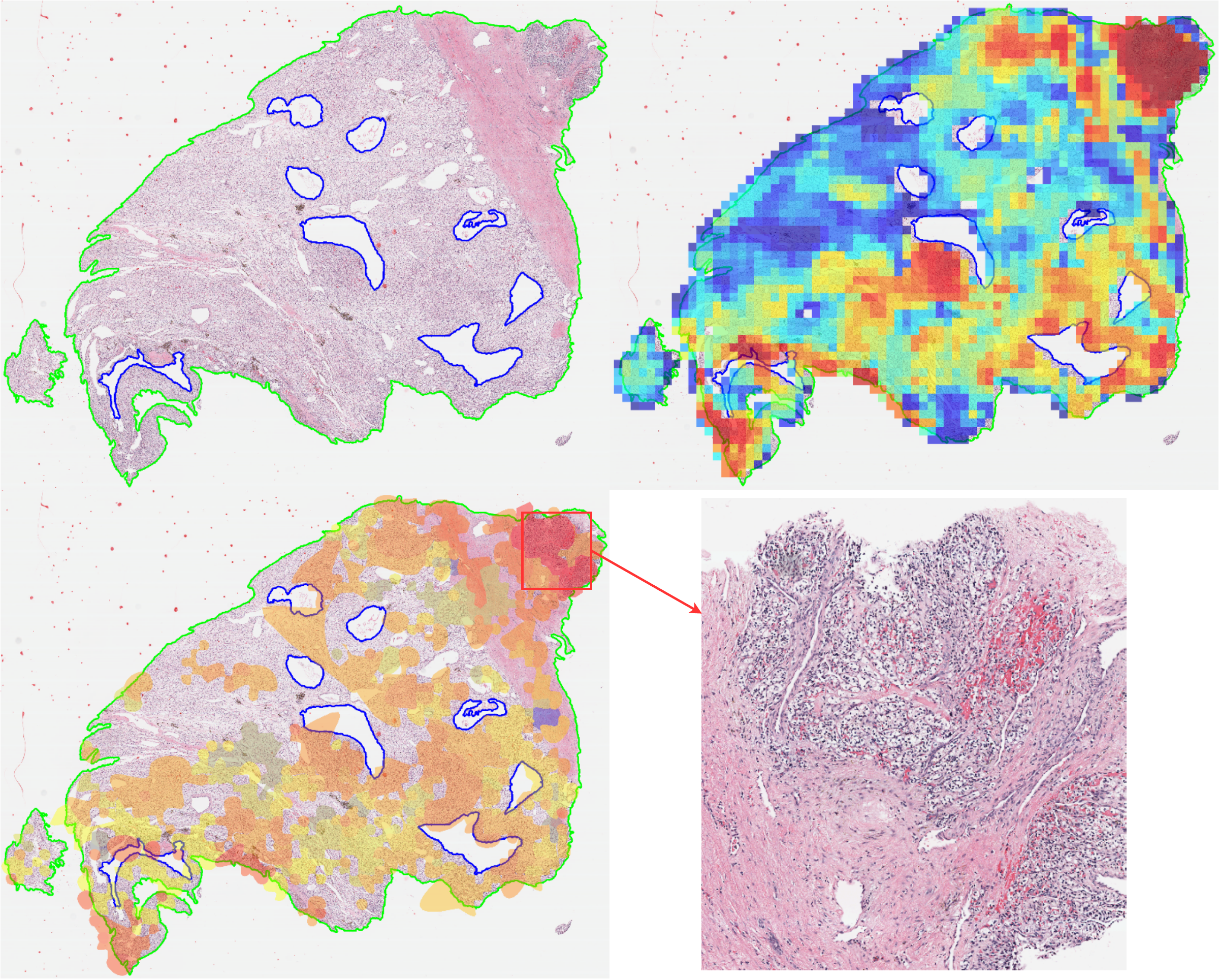}
    \captionof{figure}{Interpretability of the SCMIL.}
    \label{fig:interpret_scmil}
  \end{minipage}
   \begin{minipage}[c]{0.49\linewidth}
    \centering
    \includegraphics[width=\linewidth]{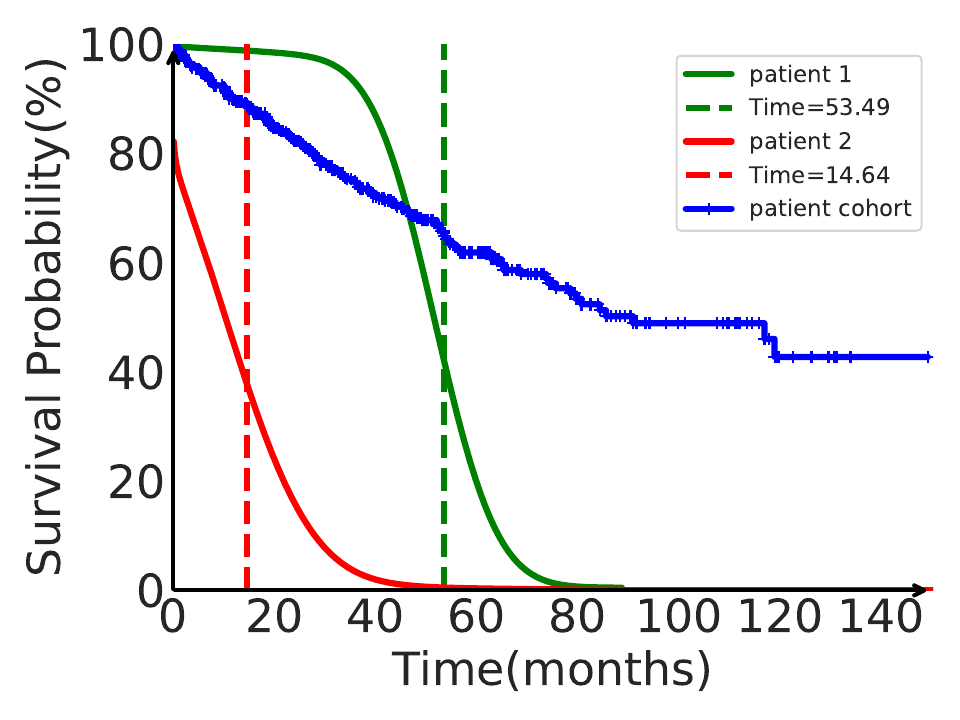}
    \captionof{figure}{Survival probability distribution prediction and actual survival time.}
    \label{fig:interpret_prediction}
  \end{minipage}
\end{figure}
\subsection{Interpretability of the Proposed Method}
We conduct an interpretability analysis for each module of SCMIL, and the visualization results are presented in Figure \ref{fig:interpret_scmil}.
The original image is located in the top left, the heatmap of IS is in the top right.
The cluster distribution image is in the bottom left, and a zoomed-in view is in the bottom right.
In the IS heat map, the color spectrum from red to yellow to blue represents a decrease in $IS$ value.
Areas closer to red are considered more valuable for the task.
In the cluster distribution image, task-relevant patches are divided into different clusters by the model, with each color representing a different cluster.
To determine which areas the model primarily focuses on for patch interactions, we calculate the average $IS$ value for patches within clusters.
The image in the bottom right of Figure \ref{fig:interpret_scmil} is an enlarged view of the region containing the cluster with the highest average $IS$ value.
The figure reveals that  the model pays more attention to the perivascular area.
Concurrently, clinical studies have identified angiogenesis and blood vessel invasion as significant factors in predicting cancer risk \cite{d2019biomarkers,kato2003combination}.
The knowledge acquired by our model coincides with clinical findings.
Figure \ref{fig:interpret_prediction} illustrates the actual survival time of two patients and the survival probability distribution predicted by our model.
The blue curve is the Kaplan-Meier curve of the patient cohort.
Our model can estimate the survival probability of patients at any given time and accurately distinguish between patients with varying survival risks.

\section{Conclusion}
In this paper, we propose SCMIL, a method designed to effectively identify instances related to survival risks from numerous instances and to discern the interactions among instances within the regions of interest.
Moreover, our method synthesizes the information from cancer patient cohort to predict a more clinically meaningful survival probability distribution for individual patient.
Experimental results on two public WSI datasets demonstrate that our method achieves superior performance and richer interpretability compared to existing methods.
In the future, we will extend our model for tasks such as predicting cancer recurrence and enhance the efficiency of our model.

\begin{credits}
\subsubsection{\ackname}This work is supported by the National Natural Science Foundation of China (62276250), the National Key R\&D Program of China (2022YFF1203303).
\subsubsection{\discintname}We have no competing interests to declare.
\end{credits}
%
% ---- Bibliography ----
%
% BibTeX users should specify bibliography style 'splncs04'.
% References will then be sorted and formatted in the correct style.
%
\bibliographystyle{splncs04}
\bibliography{Paper-2991}

\begin{thebibliography}{10}
\providecommand{\url}[1]{\texttt{#1}}
\providecommand{\urlprefix}{URL }
\providecommand{\doi}[1]{https://doi.org/#1}

\bibitem{hipt}
Chen, R.J., Chen, C., Li, Y., Chen, T.Y., Trister, A.D., Krishnan, R.G., Mahmood, F.: Scaling vision transformers to gigapixel images via hierarchical self-supervised learning. In: Proceedings of the IEEE/CVF Conference on Computer Vision and Pattern Recognition. pp. 16144--16155 (2022)

\bibitem{patchgcn}
Chen, R.J., Lu, M.Y., Shaban, M., Chen, C., Chen, T.Y., Williamson, D.F., Mahmood, F.: Whole slide images are 2d point clouds: Context-aware survival prediction using patch-based graph convolutional networks. In: Medical Image Computing and Computer Assisted Intervention--MICCAI 2021: 24th International Conference, Strasbourg, France, September 27--October 1, 2021, Proceedings, Part VIII 24. pp. 339--349. Springer (2021)

\bibitem{chen2022pan}
Chen, R.J., Lu, M.Y., Williamson, D.F., Chen, T.Y., Lipkova, J., Noor, Z., Shaban, M., Shady, M., Williams, M., Joo, B., et~al.: Pan-cancer integrative histology-genomic analysis via multimodal deep learning. Cancer Cell  \textbf{40}(8),  865--878 (2022)

\bibitem{d2019biomarkers}
D'Aniello, C., Berretta, M., Cavaliere, C., Rossetti, S., Facchini, B.A., Iovane, G., Mollo, G., Capasso, M., Pepa, C.D., Pesce, L., et~al.: Biomarkers of prognosis and efficacy of anti-angiogenic therapy in metastatic clear cell renal cancer. Frontiers in oncology  \textbf{9}, ~1400 (2019)

\bibitem{vit}
Dosovitskiy, A., Beyer, L., Kolesnikov, A., Weissenborn, D., Zhai, X., Unterthiner, T., Dehghani, M., Minderer, M., Heigold, G., Gelly, S., et~al.: An image is worth 16x16 words: Transformers for image recognition at scale. arXiv preprint arXiv:2010.11929  (2020)

\bibitem{haider2020effective}
Haider, H., Hoehn, B., Davis, S., Greiner, R.: Effective ways to build and evaluate individual survival distributions. The Journal of Machine Learning Research  \textbf{21}(1),  3289--3351 (2020)

\bibitem{hamilton2017inductive}
Hamilton, W., Ying, Z., Leskovec, J.: Inductive representation learning on large graphs. Advances in neural information processing systems  \textbf{30} (2017)

\bibitem{han2022survival}
Han, X., Goldstein, M., Ranganath, R.: Survival mixture density networks. In: Machine Learning for Healthcare Conference. pp. 224--248. PMLR (2022)

\bibitem{he2016deep}
He, K., Zhang, X., Ren, S., Sun, J.: Deep residual learning for image recognition. In: Proceedings of the IEEE conference on computer vision and pattern recognition. pp. 770--778 (2016)

\bibitem{hgt}
Hou, W., He, Y., Yao, B., Yu, L., Yu, R., Gao, F., Wang, L.: Multi-scope analysis driven hierarchical graph transformer for whole slide image based cancer survival prediction. In: International Conference on Medical Image Computing and Computer-Assisted Intervention. pp. 745--754. Springer (2023)

\bibitem{amil}
Ilse, M., Tomczak, J., Welling, M.: Attention-based deep multiple instance learning. In: International conference on machine learning. pp. 2127--2136. PMLR (2018)

\bibitem{kang2023benchmarking}
Kang, M., Song, H., Park, S., Yoo, D., Pereira, S.: Benchmarking self-supervised learning on diverse pathology datasets. In: Proceedings of the IEEE/CVF Conference on Computer Vision and Pattern Recognition. pp. 3344--3354 (2023)

\bibitem{kato2003combination}
Kato, T., Kameoka, S., Kimura, T., Nishikawa, T., Kobayashi, M.: The combination of angiogenesis and blood vessel invasion as a prognostic indicator in primary breast cancer. British journal of cancer  \textbf{88}(12),  1900--1908 (2003)

\bibitem{gcn}
Kipf, T.N., Welling, M.: Semi-supervised classification with graph convolutional networks. arXiv preprint arXiv:1609.02907  (2016)

\bibitem{dsmil}
Li, B., Li, Y., Eliceiri, K.W.: Dual-stream multiple instance learning network for whole slide image classification with self-supervised contrastive learning. In: Proceedings of the IEEE/CVF conference on computer vision and pattern recognition. pp. 14318--14328 (2021)

\bibitem{clam}
Lu, M.Y., Williamson, D.F., Chen, T.Y., Chen, R.J., Barbieri, M., Mahmood, F.: Data-efficient and weakly supervised computational pathology on whole-slide images. Nature biomedical engineering  \textbf{5}(6),  555--570 (2021)

\bibitem{raschka2020machine}
Raschka, S., Patterson, J., Nolet, C.: Machine learning in python: Main developments and technology trends in data science, machine learning, and artificial intelligence. arXiv preprint arXiv:2002.04803  (2020)

\bibitem{transmil}
Shao, Z., Bian, H., Chen, Y., Wang, Y., Zhang, J., Ji, X., et~al.: Transmil: Transformer based correlated multiple instance learning for whole slide image classification. Advances in neural information processing systems  \textbf{34},  2136--2147 (2021)

\bibitem{vaswani2017attention}
Vaswani, A., Shazeer, N., Parmar, N., Uszkoreit, J., Jones, L., Gomez, A.N., Kaiser, {\L}., Polosukhin, I.: Attention is all you need. Advances in neural information processing systems  \textbf{30} (2017)

\bibitem{wang2019machine}
Wang, P., Li, Y., Reddy, C.K.: Machine learning for survival analysis: A survey. ACM Computing Surveys (CSUR)  \textbf{51}(6),  1--36 (2019)

\bibitem{xiong2021nystromformer}
Xiong, Y., Zeng, Z., Chakraborty, R., Tan, M., Fung, G., Li, Y., Singh, V.: Nystr{\"o}mformer: A nystr{\"o}m-based algorithm for approximating self-attention. In: Proceedings of the AAAI Conference on Artificial Intelligence. vol.~35, pp. 14138--14148 (2021)

\bibitem{xu2018powerful}
Xu, K., Hu, W., Leskovec, J., Jegelka, S.: How powerful are graph neural networks? arXiv preprint arXiv:1810.00826  (2018)

\bibitem{yao2019deep}
Yao, J., Zhu, X., Huang, J.: Deep multi-instance learning for survival prediction from whole slide images. In: Medical Image Computing and Computer Assisted Intervention--MICCAI 2019: 22nd International Conference, Shenzhen, China, October 13--17, 2019, Proceedings, Part I 22. pp. 496--504. Springer (2019)

\bibitem{yao2020whole}
Yao, J., Zhu, X., Jonnagaddala, J., Hawkins, N., Huang, J.: Whole slide images based cancer survival prediction using attention guided deep multiple instance learning networks. Medical Image Analysis  \textbf{65},  101789 (2020)

\bibitem{zhu2017wsisa}
Zhu, X., Yao, J., Zhu, F., Huang, J.: Wsisa: Making survival prediction from whole slide histopathological images. In: Proceedings of the IEEE conference on computer vision and pattern recognition. pp. 7234--7242 (2017)

\end{thebibliography}

\end{document}